\documentclass{article}
\usepackage{graphicx} 
\usepackage{color}
\usepackage{booktabs}
\usepackage{multirow}
\usepackage[table,xcdraw]{xcolor}
\usepackage{amssymb}
\usepackage{xspace}
\usepackage{listings}
\usepackage{url}

\newcommand{\bench}{BRACEval\xspace}

\title{
Sabiá-3 Technical Report 
}
\author{Hugo Abonizio, Thales Sales Almeida, Thiago Laitz, \\
Roseval Malaquias Junior, Giovana Kerche Bonás,\\
Rodrigo Nogueira and Ramon Pires\\
\\
Maritaca AI}
\date{
}

\begin{document}

\maketitle

\begin{abstract}   

This report presents Sabiá-3, our new flagship language model, and Sabiazinho-3, a more cost-effective sibling. The models were trained on a large brazilian-centric corpus. Evaluations across diverse professional and academic benchmarks show a strong performance on Portuguese and Brazil-related tasks. 
Sabiá-3 shows large improvements in comparison to our previous best of model, Sabia-2 Medium, especially in reasoning-intensive tasks.
Notably, Sabiá-3's average performance matches frontier LLMs, while it is offered at a three to four times lower cost per token, reinforcing the benefits of domain specialization. 

\end{abstract}


\section{Introduction}
\label{sec:introduction}


This technical report presents the details of the development and evaluation of the Sabiá-3 and Sabiazinho-3 models. We trained them on a large corpus of documents written in Portuguese, with a special focus on Brazil-related resources. Through training, models were exposed to information relevant to Brazilian culture, history, and context. The main objective was to have a specialized model that is aware of the linguistic nuances, societal norms, and regional variations unique to the country. Throughout this report, we show that this specialization allows the models to perform better in knowledge-intensive tasks.

We applied an approach of continual learning by leveraging a ``generalist'' model that already acquired some level of language understanding and reasoning abilities, and then further trained it on our corpus of high-quality data relevant to the Brazilian context. The development consisted of two main phases: \textbf{(1)} the pre-training phase, in which we further train a pre-trained model on specialized data following a self-supervised learning strategy optimizing for the next token prediction objective, and \textbf{(2)} the post-training phase where the model is tuned to follow instructions and align to human preferences.

Compared to our previous release, Sabiá-2~\cite{almeida2024sabia2}, we have collected a significantly larger volume of data for pre-training. In addition to the scale, we also improved the quality of our pre-training data by using a mixture of heuristic and model-based methods to filter out low-quality data.

Considering the challenges of training large scale models, we used TPU v5 accelerators with Jax~\cite{jax2018github} to perform distributed training. We combined both data and model parallelism to achieve high training throughput and to ensure efficient utilization of TPU hardware.

After pre-training, we employed a mix of human-annotated and synthetically generated data to teach our base model to follow instructions~\cite{ai2024yi,phi3,gemini,gemma,nemotron,llama3}. The instruction tuning data is composed of both human-annotated prompts and synthetically generated instructions targeting specific capabilities. In addition, we also include synthetic examples to induct self-awareness abilities in the model~\cite{takenoutofcontext}.

\begin{figure*}[t]
    \centering
    \includegraphics[width=\textwidth]{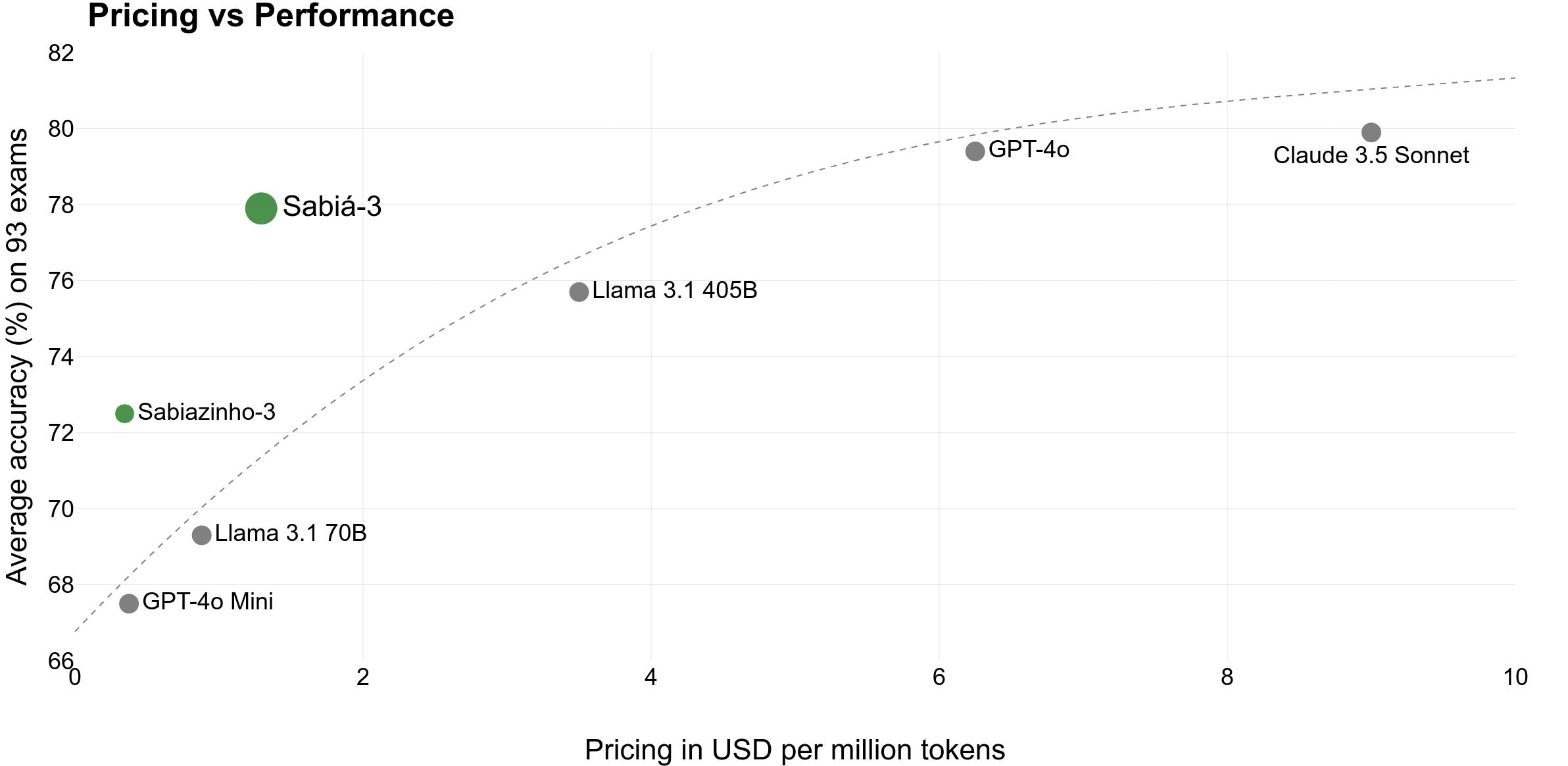}
    \caption{Price in USD per million tokens, considering an equal proportion of input and output tokens, versus performance on 70 Brazilian exams (in Portuguese). The dashed curve represents a Pareto frontier for general-purpose LLMs such as Llama 3.1 and GPT-4o, which Sabiá-3 and Sabiazinho-3 surpass due to their domain specialization.}
    \label{fig:pareto}
\end{figure*}







\section{Evaluation}
\label{sec:evaluation}

In this section, we compare Sabiá-3 and Sabiazinho-3 with proprietary and open-source LLMs, listed in Table~\ref{table:model-versions}, on various benchmarks.

\begin{table}[h]
\centering
\resizebox{0.9\textwidth}{!}{
\begin{tabular}{l|c|c|rr}
\toprule
& & \textbf{Knowledge} & \multicolumn{2}{c}{\textbf{USD per 1M tokens}}\\ 
\textbf{Model} & \textbf{Version} & \textbf{Cutoff} & \textbf{Input} & \textbf{Output}\\
\midrule
Sabiá-3 & 2024-09-09 & Mid-2023 & 0.83 & 1.67 \\
Sabiazinho-3 & 2025-02-06 & Mid-2023 & 0.17 & 0.50 \\
Sabiá-2 Medium & 2024-03-13 & Mid-2023 & \multicolumn{2}{c}{Deprecated} \\
GPT-4o & 2024-08-06 & October 2023 & 2.50 & 10.00 \\
GPT-4o Mini & 2024-07-18 & October 2023 & 0.15 & 0.60 \\
Claude 3.5 Sonnet & 2024-06-20 & April 2024 & 3.00 & 15.00 \\
Llama 3.1 8B  & Instruct & December 2023 & 0.18 & 0.18 \\
Llama 3.1 70B & Instruct & December 2023 & 0.88 & 0.88 \\
Llama 3.1 405B & Instruct & December 2023 & 3.50 & 3.50 \\
\bottomrule
\end{tabular}
}
\caption{Overview of the models evaluated in this work, with pricing information as of October 14, 2024. The API pricing for Llama 3.1 is from together.ai.}
\label{table:model-versions}
\end{table}

\subsection{Multiple-choice Exams}
\label{sec:exams}

We have developed a benchmark suite to evaluate the Sabiá-3 performance on 93 academic exams, including entry-level, undergraduate, and professional certification exams. Our focus is on multiple-choice questions from various Brazilian educational assessments, particularly those administered after the middle of 2023 to mitigate the risk of data contamination. However, ENADE is an exception, including spans of 2022 and 2023 to cover a wide range of disciplines.

For more details about the datasets, we recommend the readers to the Sabiá-2 technical report~\cite{almeida2024sabia2}.
In addition to the 8 benchmarks introduced in~\cite{almeida2024sabia2}, we now incorporated five new multiple-choice benchmarks:

\begin{itemize}
    \item ENAM: The examination aims to qualify law graduates interested in participating in magistracy selection processes promoted by federal regional courts, labor courts, military courts, and courts of the states and the Federal District and territories. The benchmark covers the first edition of the exam, administered on April 14, 2024, and its subsequent re-application one month later with a different set of questions. There are 160 questions in total.
    \item CPNU: First edition of the \textit{Concurso Público Nacional Unificado} (Unified National Public Competition), administered by the Ministry of Management and Innovation in Public Services on August 18, 2024 for filling permanent public positions. The benchmark includes eight exams with a total of 370 questions, covering the following thematic areas:
    \begin{enumerate}
        \item Infrastructure, Exact Sciences, and Engineering;
        \item Technology, Data, and Information;
        \item Environmental, Agricultural, and Biological Sciences;
        \item Work and Public Servant Health;
        \item Education, Health, Social Development, and Human Rights;
        \item Economic Sectors and Regulation;
        \item Government Management and Public Administration; and
        \item Intermediate Level.
    \end{enumerate}
    \item BCB: The BCB benchmark refers to the public examination for the Central Bank of Brazil. This exam features objective questions with two possible answers: true or false. It comprises three distinct exams: basic knowledge, information technology, and economics and finance.
    \item BNDES: The BNDES benchmark is the public examination for the National Bank for Economic and Social Development. Similar to the BCB exam, it consists of objective questions with two alternatives: true or false. The benchmark includes 14 exams covering a range of subjects such as law, systems analysis, administration, architecture, economics, engineering, accounting, and psychology.
    \item CFCEQ: The CFCEQ benchmark is the technical qualification exam conducted by the Federal Accounting Council (CFC). Its purpose is to assess the knowledge and professional competence of accountants who wish to work as independent auditors in organizations regulated by the CVM, BCB, Susep, and Previc. The benchmark comprises 12 exams.

\end{itemize}

\begin{figure*}[t]
    \centering
    \includegraphics[width=\textwidth]{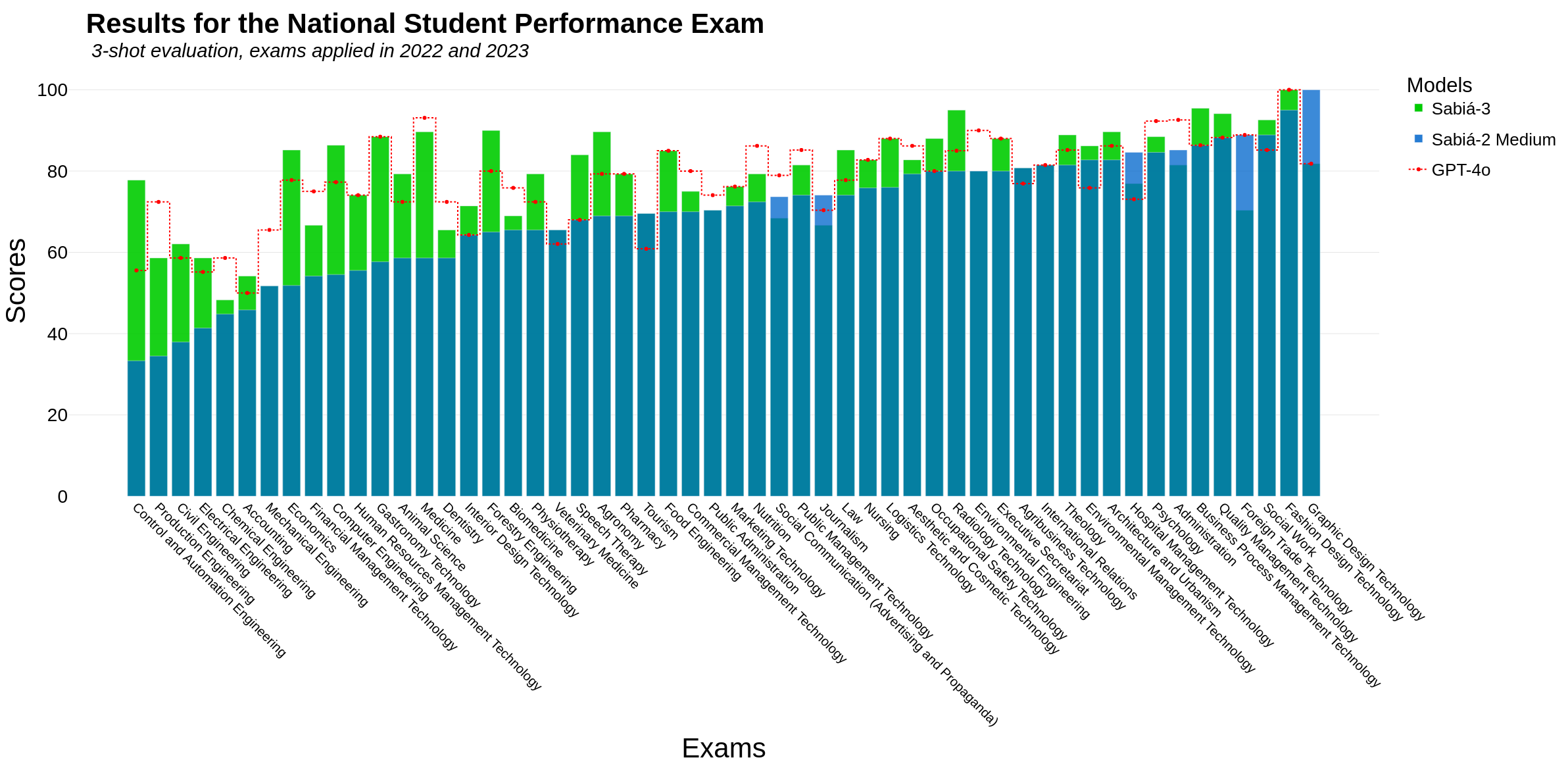}
    \caption{Accuracies of Sabiá-3, Sabiá-2 Medium and GPT-4o on Enade 2022 and 2023 exams, ordered from low to high based on Sabiá-2 Medium performance. Sabiá-3 outperforms Sabiá-2 Medium on 76\% of the exams.} 
    \label{fig:enade}
\end{figure*}

Figure~\ref{fig:enade} compares the performance of Sabiá-3, Sabiá-2 Medium, and GPT-4o on the ENADE 2022 and 2023 exams. 
The results reveal that Sabiá-2 has a higher difficulty level in engineering and economics-related areas, as indicated by its lowest 10 accuracies. In contrast, Sabiá-3 shows a significant improvement in these challenging areas, achieving a 70\% reduction in errors compared to Sabiá-2, particularly in Economics and Computer Engineering. 
Moreover, Sabiá-3 demonstrates a competitive performance compared with GPT-4o, performing equally well or superior in 36 of the 54 Enade exams.
Despite these advancements, the results indicate that Sabiá-3 continues to face difficulties in certain areas, notably Chemical Engineering, Mechanical Engineering, and Accounting, where it reached its lowest scores.

Table~\ref{tab:results_per_benchmark_discipline} shows the scores of the models in 13 benchmarks derived from academic and professional exams applied after mid-2023. The results illustrate that Sabiá-3 has a comparable level of accuracy to GPT-4o across all 93 exams assessed. GPT-4o exhibits an advantage particularly in the Medicine benchmarks, while Sabiá-3 demonstrates a notable performance in the CPNU exams. Furthermore, Sabiá-3 surpasses Llama-3.1 405B by 2 percentage points in accuracy, performing equally or better in 62\% of the exams. Claude 3.5 Sonnet achieves the highest average accuracy, although it presents a higher risk of data contamination due to its training data cutoff date.

\begin{table}
\centering
\resizebox{\textwidth}{!}{ 
\begin{tabular}{lcc|cc|cc|c|ccc}
\toprule
\textbf{} & \textbf{} & \textbf{} & \textbf{Llama-3.1} & \textbf{Llama-3.1} & \textbf{GPT-4o} & \textbf{GPT-4o} & \textbf{Claude 3.5} & \textbf{Sabiá-2} & \textbf{Sabiazinho-3} & \textbf{Sabiá-3} \\
\textbf{} & \textbf{Discipline} & \textbf{Exams} & \textbf{70B} & \textbf{405B} & \textbf{Mini} & \textbf{} & \textbf{Sonnet} & \textbf{Medium} & \textbf{} & \textbf{} \\
\midrule
\textbf{ENEM} & Multiple        & 2 & 83.8\% & 84.9\% & 83.8\% & 87.2\% & 89.9\% & 71.8\% & 82.7\% & 87.7\% \\
\textbf{BLUEX} & Multiple       & 2 & 77.3\% & 83.9\% & 85.1\% & 88.7\% & 93.0\% & 75.9\% & 83.7\% & 86.2\% \\
\textbf{ENADE} & Multiple       & 28 & 66.2\% & 73.0\% & 63.2\% & 75.2\% & 78.0\% & 64.6\% & 71.9\% & 77.2\% \\
\textbf{CPNU} & Multiple        & 8 & 82.2\% & 88.9\% & 83.4\% & 88.5\% & 83.8\% & 79.2\% & 85.5\% & 90.6\% \\
\textbf{BNDES} & Multiple       & 14 & 76.5\% & 80.0\% & 76.5\% & 86.7\% & 81.2\% & 75.5\% & 80.6\% & 84.3\% \\
\textbf{BCB} & Multiple         & 3 & 75.8\% & 79.4\% & 76.3\% & 84.1\% & 81.8\% & 65.0\% & 76.5\% & 80.1\% \\
\textbf{POSCOMP} & Comp. Sci    & 1 & 61.5\% & 78.5\% & 63.1\% & 73.8\% & 76.9\% & 50.8\% & 76.9\% & 76.9\% \\
\textbf{OAB} & Law              & 5 & 68.6\% & 77.6\% & 64.1\% & 83.4\% & 81.9\% & 72.1\% & 72.6\% & 76.4\% \\
\textbf{ENAM} & Law             & 3 & 53.8\% & 59.3\% & 50.0\% & 70.2\% & 75.2\% & 58.9\% & 62.6\% & 70.6\% \\
\textbf{CFCES} & Account.       & 4 & 61.3\% & 68.9\% & 65.3\% & 75.4\% & 80.4\% & 59.3\% & 70.4\% & 78.4\% \\
\textbf{CFCEQ} & Account.       & 12 & 61.3\% & 72.1\% & 58.3\% & 73.5\% & 79.2\% & 63.0\% & 66.7\% & 70.7\% \\
\textbf{Revalida} & Medicine    & 3 & 81.3\% & 84.2\% & 74.8\% & 87.1\% & 85.7\% & 73.0\% & 76.2\% & 83.5\% \\
\textbf{MREX} & Medicine        & 8 & 65.5\% & 69.0\% & 60.9\% & 76.1\% & 73.7\% & 53.4\% & 52.4\% & 63.7\% \\

\midrule
\textbf{Average} & -            & 93 & 70.4\% & 76.9\% & 69.6\% & 80.8\% & 81.6\% & 66.4\% & 73.7\% & 79.0\% \\
\bottomrule
\end{tabular}
}
\caption{Accuracy on academic benchmarks. The average results (last row) cover all the 93 exams applied after the middle of 2023, which is the knowledge cutoff date of Sabiá-3. Claude 3.5's cutoff date is April of 2024, so exams released before that date might be present in its training data.}
\label{tab:results_per_benchmark_discipline}
\end{table}

In Figure~\ref{fig:pareto}, we show the cost per token as of October 2024 in comparison to the average accuracy on the aforementioned exams. We have plotted a Pareto curve based on generalist models, i.e., the ones not specialized in the Brazilian context. Sabiá-3 and Sabiazinho-3 stay above this frontier, underscoring the cost-quality benefits that result from domain specialization.

\subsection{Conversation Capabilities}
\label{sec:conv}

This section presents the benchmark we use to evaluate the Sabiá-3 capabilities to engage in dialogues.

In our previous work \cite{almeida2024sabia2}, we introduced the Brazilian Chat Evaluation (\bench), a benchmark inspired on MT-Bench~\cite{zheng2024judging} and designed to evaluate AI assistants' performance in following instructions, engaging in multi-turn dialogues, and demonstrating knowledge specific to Brazil. \bench consists of 150 multi-turn questions distributed across 13 categories, including culturally relevant topics such as writing, roleplay, extraction, humanities, entity, and contradiction analysis, as well as universal categories like abstention, harmful content, reasoning, math, and coding.

\begin{figure*}[!htbp]
    \centering
    \includegraphics[width=0.8\textwidth]{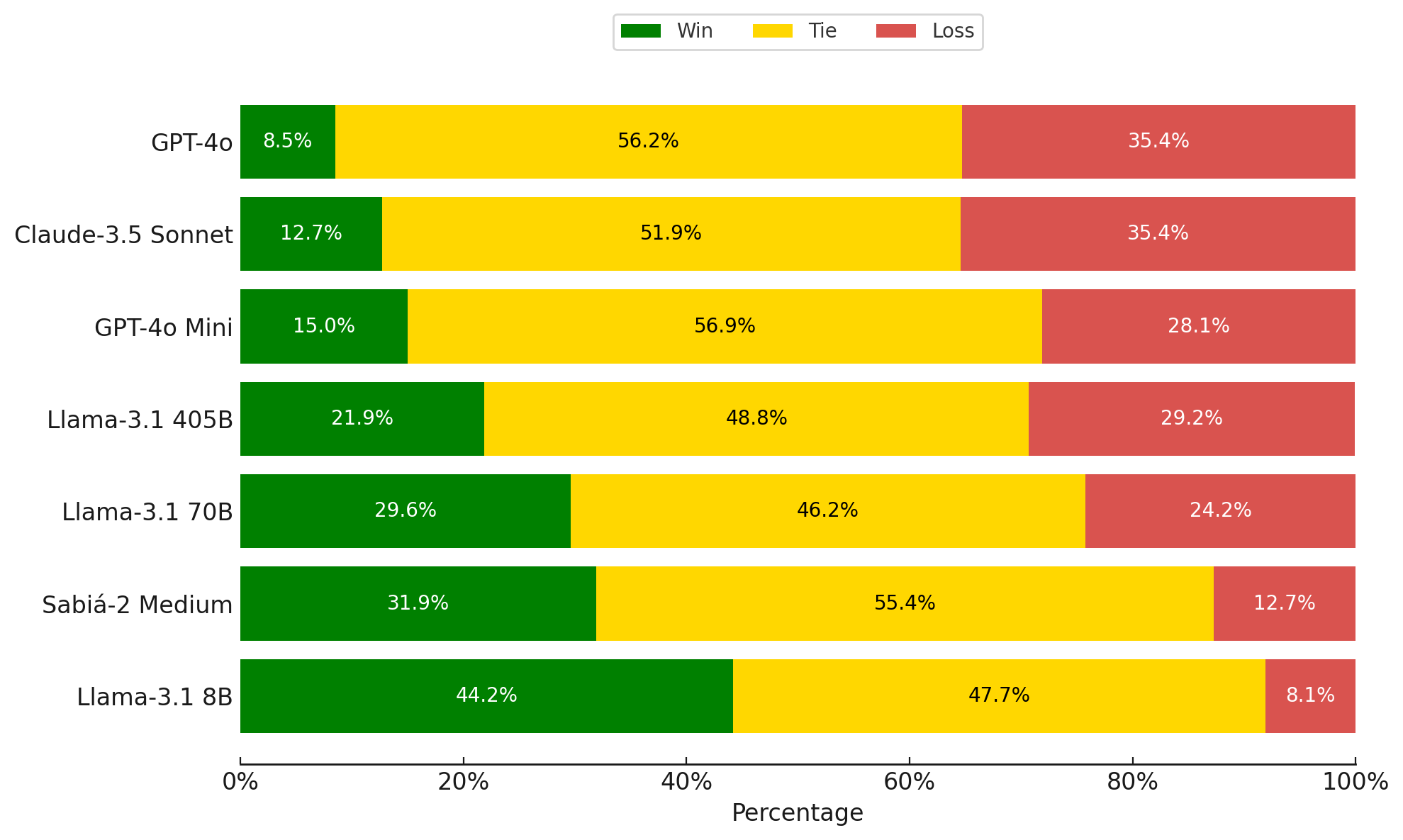}
    \caption{Win, tie and loss rates for Sabiá-3 against other LLMs on the \bench conversation benchmark according to GPT-4-turbo as a judge.}
    \label{fig:results_braceval}
\end{figure*}

Figure~\ref{fig:results_braceval} compares Sabiá-3 with other LLMs on \bench. The judge is an LLM that is given the prompt, the answer from Sabiá-3 and another competitor model and decides which one won or a tie. We address position bias by calling the judge twice with the order of two answers swapped.

We use GPT-4-turbo (more specifically, gpt-4-1106-preview) as the judge because, in pairwise evaluations against other models like Claude 3.5 Sonnet and GPT-4o, all judges — including Claude 3.5 Sonnet, GPT-4-turbo, and GPT-4o — consistently preferred the responses from GPT-4-turbo.

Sabiá-3 shows a significant improvement over Sabiá-2 Medium, despite still having a high percentage of losses against more expensive models, such as GPT-4o and Claude 3.5 Sonnet. A noteworthy result is GPT-4o Mini, which, despite its lower per-token price, achieves a higher win rate compared to the more expensive Llama-3.1 405B in this benchmark.

\begin{figure*}[!htbp]
    \centering
    \includegraphics[width=\textwidth]{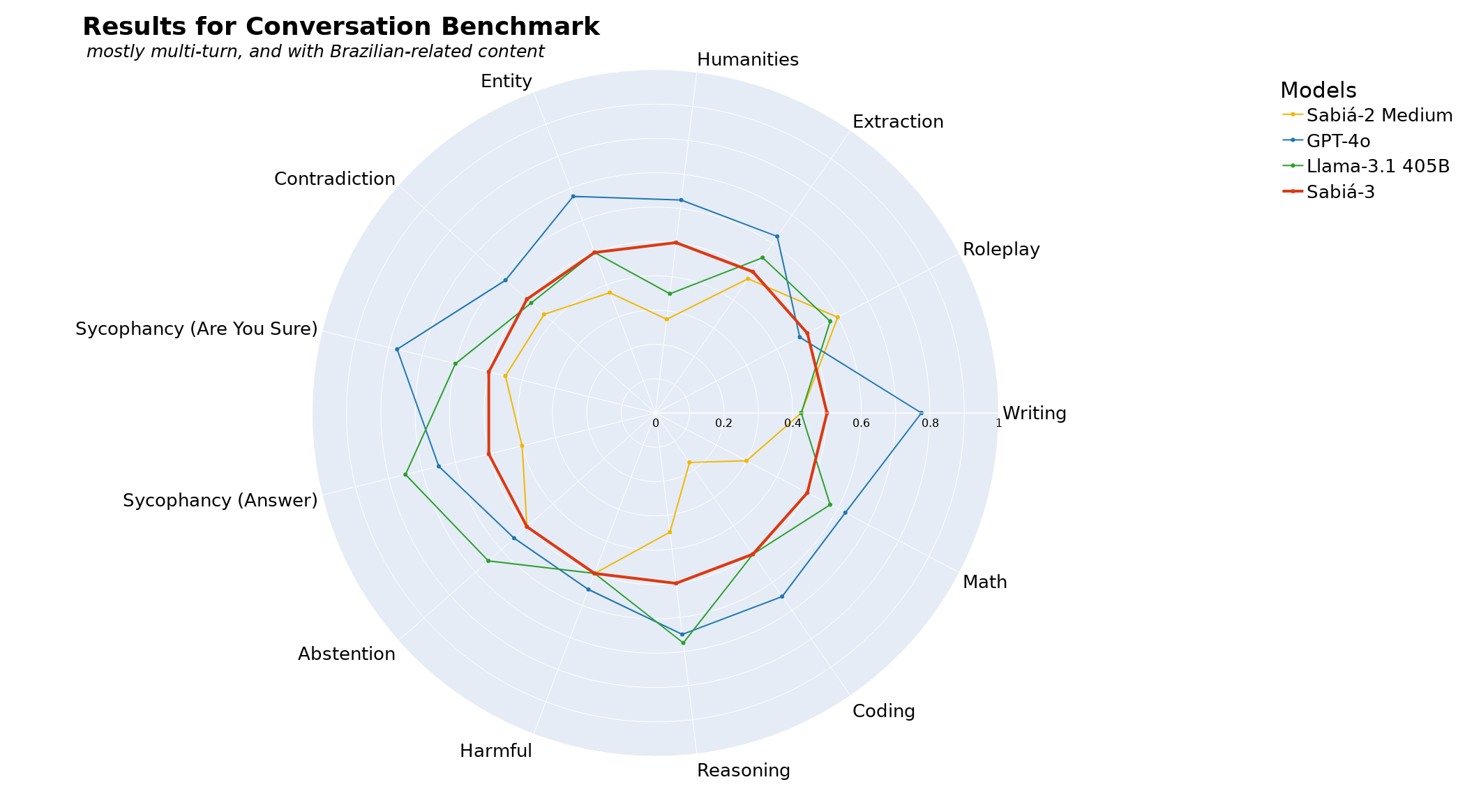}
    \caption{Category-wise adjusted win rates of Sabiá-3 against other models on \bench, with 0.5 representing a tie; models that score below it are worse than Sabiá-3; above it, competitor models are superior. 
    }
    \label{fig:results_conversation_bench_categories}
\end{figure*}

Figure~\ref{fig:results_conversation_bench_categories} provides a visual representation of the performance of Sabiá-3 relative to other models, across each \bench category. We measure the performance using the adjusted win rate, where the winner model receives one point, and in the case of a tie, both models earn 0.5 points each. 
Across most categories, Sabiá-3 outperforms Sabiá-2 Medium and performs comparably to the more expensive Llama-3.1 405B model. However, it performs poorly compared to GPT-4o, particularly in coding, math, and writing. This result is consistent regardless of the judge model; using Claude 3.5 Sonnet as a judge yields similar win rates, which indicates that the result is not influenced by the use of GPT-4-turbo as the judge.



\subsection{Instruction-Following Capabilities}

One relevant benchmark to assess LLM abilities to follow instructions is the IFEval~\cite{zhou2023instruction}. IFEval consists of approximately 500 prompts including instructions that are veriafiable programmatically such as ``write in more than 400 words". It uses two main metrics: strict and loose. In the strict metric, the model's response is checked against a set of verifiable instructions to determine if it followed all of them. If so, the model receives a point. If any part of the instruction is not followed exactly as specified, the response is deemed incorrect and the model does not receive a point. The loose metric is more lenient as it considers variations and common transformations that might occur in the model's response. For example, if an instruction requires a specific phrase to be included at the end of an email, but the model includes that phrase with additional text formatting like bold or italics, the loose metric might still consider the instruction to be followed, whereas the strict metric would not.

Since this benchmark's prompts and target answers are in English, we do not expect our Portuguese-specialized model to yield significant improvements in this scenario. Instead, this benchmark primarily serves to assess how well our model can follow instructions in comparison to state-of-the-art LLMs.

The results presented in Table~\ref{table:ifeval} show that Sabiá-3 and Sabiazinho-3 outperform Sabiá-2 Medium in terms of instruction-following capabilities. However, it lags behind other models evaluated in this study, including cheaper options such as GPT-4o Mini and Llama-3.1 8B.

\begin{table}[h]
\centering
\begin{tabular}{lcc}
\toprule
\textbf{Model} & \textbf{Loose} & \textbf{Strict} \\
\midrule
GPT-4o & 88.4 & 83.7 \\
GPT-4o Mini & 83.9 & 81.0 \\
\midrule
Llama-3.1 405B  & 88.0 & 85.6 \\
Llama-3.1 70B  & 87.9 & 83.1 \\
Llama-3.1 8B  & 80.5 & 76.6 \\
\midrule
Sabiá-3 & 78.0 & 73.8 \\
Sabiazinho-3 & 81.5 & 77.3 \\
Sabiá-2 Medium & 41.8 & 38.6 \\
\bottomrule
\end{tabular}
\caption{Results on the IFeval benchmark.}
\label{table:ifeval}
\end{table}

\subsection{Long Context Capabilities}
\label{sec:niah}

Sabiá-3 and Sabiazinho-3 can process up to 32,000 tokens in a sequence. Here we evaluate how well the model uses this capacity through the Needle-in-the-Haystack (NIAH) benchmark~\cite{NIAH}, which measures the model's ability to find answers within a long, unrelated context. The test involves inserting a small piece of text within a larger context ranging from 1,000 to 32,000 tokens, at various depths from 0\% to 100\%. Since we are focusing on evaluating the performance of an LLM on Portuguese tasks, we adapt this benchmark using the book \textit{Dom Casmurro}\footnote{\url{https://machadodeassis.net/texto/dom-casmurro/11503}} as the context. In a random spot within the text, we insert as a needle a sentence that says ``\textit{O número mágico de Campinas é} \{random\_number\}.'' (``The magic number of Campinas is \{random\_number\}.''), replacing \{random\_number\} with a 6-digit number. Then, we provide this modified context along with the question \textit{``Qual o número mágico de Campinas?''} (``What is the magic number of Campinas?'') to the model. Using regular expression, we check whether the model correctly outputs the random number as the answer.

\begin{figure*}[!htbp]
    \centering
    \includegraphics[width=\textwidth]{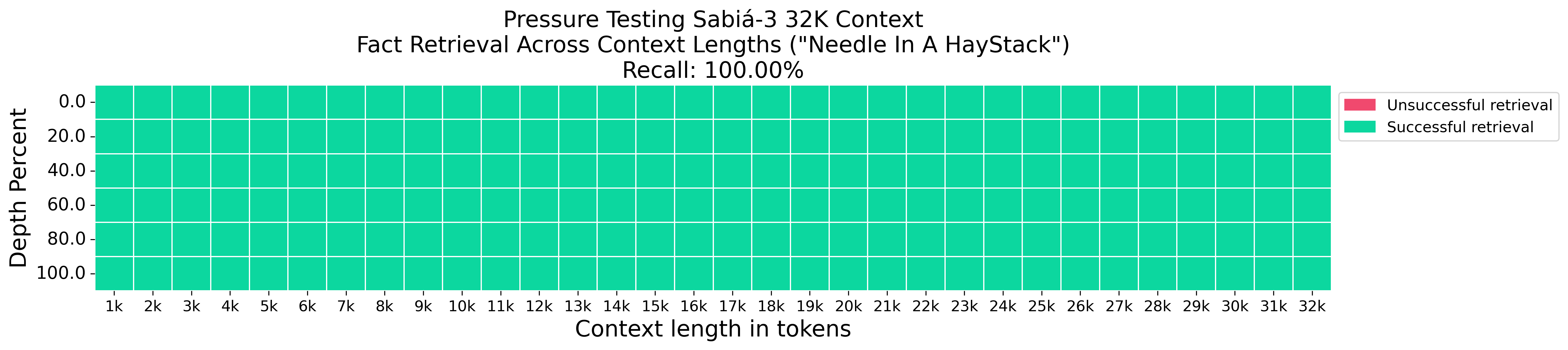}
    \caption{Performance of the Sabiá-3 model in the Portuguese-adapted Needle-in-the-Haystack (NIAH) benchmark.}
    \label{fig:results_needleinhaystack}
\end{figure*}

\begin{figure*}[!htbp]
    \centering
    \includegraphics[width=\textwidth]{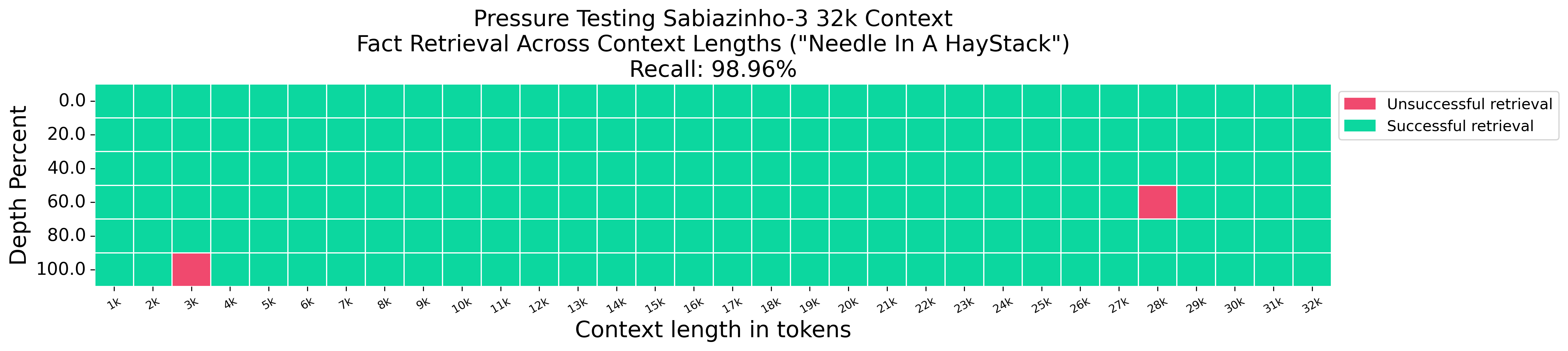}
    \caption{Performance of the Sabiazinho-3 model in the Portuguese-adapted Needle-in-the-Haystack (NIAH) benchmark.}
    \label{fig:results_needleinhaystack2}
\end{figure*}

As depicted in Figures~\ref{fig:results_needleinhaystack} and~\ref{fig:results_needleinhaystack2}, Sabiá-3 exhibited a perfect recall in our Portuguese version of the NIAH benchmark and Sabiazinho-3 reached a recall of 99.48\%, consistently locating relevant information within long contexts. We acknowledge, however, that the task, due to its simplicity, may not fully capture the long-context capabilities of language models. Therefore, for a better measure of long-range comprehension skills, we need Portuguese versions of more challenging benchmarks, such as the QuALITY benchmark~\cite{pang2022quality}, which tests a model's ability to answer questions about books.

\subsection{Function Calling Capabilities}
\label{sec:bfcl}

The Berkeley Function Calling Leaderboard (BFCL)~\cite{berkeley-function-calling-leaderboard} is a benchmark for evaluating the LLM's capabilities in executing function calls and using tools effectively. The benchmark is updated periodically, and comprises three distinct subsets:
\begin{enumerate}
    \item Non-live: An expert-curated collection of question-function-answer pairs, encompassing multiple languages and complex use cases, designed to challenge the models' ability to understand and apply functions accurately. The examples are categorized into Abstract Syntax Tree (AST) analysis, measuring whether the function names, arguments, and parameter types match the expected response; and executable verification (Exec), which executes the function and verifies whether the output is correct.
    \item Live: User-contributed examples that reflect real-world applications, ensuring that the models are tested against scenarios that mirror everyday tasks.
    \item Multi turn: Multi-turn and multi-step function calling tasks that simulate agentic behaviors, requiring models to plan execution strategies, request information, and manage a sequence of function invocations to complete a task.
\end{enumerate}

The benchmark also includes two categories to measure hallucinations: irrelevance detection, when none of the provided functions is relevant to the query and should not be invoked; and relevance, when at least one of the functions should be invoked.

To evaluate the models we use the original examples in English. Table~\ref{table:bfcl} shows the results of Sabiá-3, Sabiazinho-3, and other proprietary LLMs. 
The results indicate that Sabiá-3 ranks competitively alongside top-performing models. It outperforms in the Non-live categories but requires enhancements in multi-turn and multi-step tasks. For instance, increasing the support for more than 32,000 tokens could improve in one of the multi-turn categories. It also needs to improve in the irrelevance category.

\begin{table}[htbp]
\centering
\resizebox{\textwidth}{!}{
\begin{tabular}{@{}lccccccc@{}}
\toprule
\textbf{Model} & \textbf{Overall Acc} & \multicolumn{3}{c}{\textbf{Single Turn}} & \textbf{Multi Turn} & \multicolumn{2}{c}{\textbf{Hallucination}} \\
\cmidrule(lr){3-5} \cmidrule(lr){6-6} \cmidrule(lr){7-8}
& & \textbf{Non-live (AST)} & \textbf{Non-live (Exec)} & \textbf{Live (AST)} & \textbf{Overall Acc} & \textbf{Relevance} & \textbf{Irrelevance} \\
\midrule
GPT-4o           & 69.58\% & 87.42\% & 89.02\% & 79.65\% & 41.00\% & 83.33\% & 83.15\% \\
GPT-4o Mini      & 64.10\% & 85.21\% & 83.57\% & 74.41\% & 34.12\% & 83.33\% & 74.75\% \\
Sabiá-3          & 56.50\% & 90.65\% & 91.46\% & 75.83\% & 7.00\% & 88.89\% & 59.90\% \\
Claude 3.5 Sonnet& 56.46\% & 45.44\% & 47.89\% & 78.94\% & 41.00\% & 77.78\% & 74.04\% \\
Sabiazinho-3     & 55.21\% & 86.62\% & 82.67\% & 75.08\% & 7.25\% & 77.78\% & 74.61\% \\

\bottomrule
\end{tabular}
}
\caption{Results on the BFCL benchmark ordered by overall accuracy.}
\label{table:bfcl}
\end{table}

\subsection{Agentic Performance}
\label{sec:agent}

Recent advancements in the ability of LLMs on instruction-following, function calling, and reason in long contexts have enabled the development of LLM-based agents. These agents can automatically execute complex tasks, such as developing code \cite{zhang2024codeagent,DEVIN}, managing everyday user tasks on operating systems \cite{mei2024aiosllmagentoperating,RECALL}, and orchestrating robots \cite{ahn2024autortembodiedfoundationmodels}.

To evaluate an LLM in agentic tasks, it is necessary to integrate it with a framework with modules for memory, action, and reasoning. This setup allows the LLM to interact with responsive environments through tools using natural language. To this end, we use AgentBench \cite{liu2023agentbenchevaluatingllmsagents}, a framework designed to evaluate the agentic capabilities of LLMs across eight distinct environments. It tests the model's performance in multi-turn open-ended generation scenarios involving coding, gaming, and web.

Here are the eight environments where the agent LLM performs tasks in AgentBench:

\begin{itemize}
\item Operating System (OS): Interacts with a real Ubuntu terminal, executing bash scripts to retrieve information about the virtual OS.
\item Database (DB): Runs SQL queries on a database to answer high-level questions about multiple tables.
\item Knowledge Graph (KG): Navigates a knowledge graph with 45 million entities and 3 billion facts to extract high-level insights about relationships between entities.
\item Web Shopping (WS): Navigates an online shopping environment to make purchases from a web page.
\item Web Browsing (WB): Browses web pages across various domains to extract information or perform high-level actions.
\item Digital Card Game (DCG): Plays a game against a rule-based agent, analyzing cards with different abilities to defeat the opponent.
\item Lateral Thinking Puzzle (LTP): Solves riddles by asking questions where another agent LLM (GPT-3.5 Turbo) can only respond with ``yes’’, ``no’’, or ``irrelevant’’.
\item House-Holding (HH): Explores a virtual house, interacting with objects and performing high-level tasks.
\end{itemize}

Table \ref{table:agentbench} presents the results, using the same metrics defined in the AgentBench paper \cite{liu2023agentbenchevaluatingllmsagents}. Each environment metric ranges from 0 to 100, with higher scores indicating better performance. We report the Overall Average (OA) across all environments as our primary metric. Due to the variation in mean scores across different environments, OA is the weighted average of tasks scores, normalized to a scale from 0 to 10, where the weights are inversely proportional to the average score obtained by the models evaluated in the AgentBench paper. That is, harder tasks will have a higher weight.

\begin{table}[h]
\centering
\scalebox{0.9}{\begin{tabular}{l c c c c c c c c c}
\toprule    
\multirow{2}{*}{Models} & \multirow{2}{*}{\textbf{OA}} & \multicolumn{3}{c}{Code-grounded} & \multicolumn{2}{c}{Web-grounded} & \multicolumn{3}{c}{Game-grounded} \\
\cmidrule(lr){3-5} \cmidrule(lr){6-7} \cmidrule(lr){8-10}
&  & \textbf{OS} & \textbf{DB} & \textbf{KG} & \textbf{WS} & \textbf{WB} & \textbf{DCG} & \textbf{LTP} & \textbf{HH}\\
\midrule
GPT-4o & 3.7 & 38.9 & 52.7 & 58.3 & 60.0 & 12.0 & 73.1 & 11.9 & 72.0\\
GPT-4o Mini & 3.1 & 29.7 & 51.0 & 38.6 & 57.3 & 6.0 & 89.0 & 7.5 & 40.0\\
\midrule
Llama-3.1 405B  & 3.8 & 43.0 & 58.3 & 48.1 & 55.9 & 35.0 & 58.6 & 9.6 & 82.0\\
Llama-3.1 70B  & 3.4 & 43.0 & 56.7 & 42.9 & 44.7 & 23.0 & 78.1 & 7.2 & 54.0\\
Llama-3.1 8B  & 1.5 & 16.7 & 0.0 & 15.0 & 54.1 & 17.0 & 55.1 & 0.2 & 22.0\\
\midrule
Sabiá-3 & 2.9 & 29.2 & 28.7 & 36.7 & 62.5 & 30.0 & 57.5 & 7.6 & 58.0\\
Sabiazinho-3 & 1.9 & 20.1 & 27.3 & 15.7 & 53.57 & 23.0 & 13.78 & 5.4 & 46.0\\
Sabiá-2 Medium & 0.9 & 12.5 & 2.6 & 4.9 & 44.6 & 20.0 & 0.7 & 3.0 & 20.0\\
\bottomrule
\end{tabular}}
\caption{Results on AgentBench.}
\label{table:agentbench}
\end{table}
Even though AgentBench is in English, Sabiá-3 shows competitive performance compared to models primarily trained in this language, achieving an OA of 2.9, comparable to GPT-4o Mini. Notably, Sabiá-3 shows an improvement of 2 OA over its predecessor, Sabiá-2 Medium.

Sabiá-3 particularly excels in web-grounded agentic tasks, which involve navigating through web pages to extract relevant information and perform high-level actions. It achieves the top performance in WS, with a score of 62.5, and the second-best performance in WB. However, there's still improvement to be done in code-grounded environments such as OS and DB.



\section{Conclusion}
\label{sec:conclusion}

We introduced Sabiá-3 and Sabiazinho-3, a family of language models focused on Brazil-related content. They improve upon our earlier Sabiá-2 models, especially in handling long texts, reasoning, and coding. Despite rivaling state-of-the-art proprietary and open-source models on knowledge-intensive tasks while having a lower per token cost, Sabiá-3 and Sabiazinho-3 still have room for improvement in multi-step tasks and following instructions accurately, both of which we plan to address in future work.




\bibliographystyle{plain}
\bibliography{references}

\end{document}